\documentclass[letterpaper, 10 pt, conference]{ieeeconf}  

\IEEEoverridecommandlockouts                              

\makeatletter
\let\NAT@parse\undefined
\makeatother
\usepackage[numbers,sort&compress]{natbib}

\usepackage[pdftex]{graphicx}
\usepackage{amsmath}
\usepackage{amssymb}  
\usepackage{subfigure}
\usepackage{subcaption}
\usepackage{multirow}
\usepackage{array,booktabs}
\usepackage{diagbox}
\usepackage{balance}
\usepackage{xspace}
\usepackage{algorithm}
\usepackage{algpseudocode}
\usepackage{adjustbox}
\usepackage{romannum}
\usepackage{float}
\usepackage{placeins}

\usepackage[final]{hyperref}
\hypersetup{
 colorlinks=true,
 linkcolor=blue,
 filecolor=magenta,
 urlcolor=blue,
 citecolor=black
}
\usepackage{cleveref}
\crefname{figure}{Fig.}{Figs.}
\Crefname{figure}{Fig.}{Figs.}
\usepackage[font=small]{caption}
\usepackage{rpm_SIunits}
\usepackage{rpm_acronyms}
\usepackage{rpm_math}
\usepackage{rpm_misc}

\usepackage{soul,color}
\usepackage{lipsum}

\usepackage{xcolor}

\usepackage{tikz} 
\usepackage{subcaption}
\usepackage{caption}
\usepackage{amssymb}
\usepackage{multirow}
\usepackage{graphicx}
\title{\LARGE \bf HeRCULES: Heterogeneous Radar Dataset\\
in Complex Urban Environment for Multi-session Radar SLAM
}     

\author{Hanjun Kim${}^{1}$, Minwoo Jung${}^{2}$, Chiyun Noh${}^{2}$, Sangwoo Jung${}^{2}$,\\Hyunho Song${}^{2}$, Wooseong Yang${}^{2}$, Hyesu Jang${}^{2}$ and Ayoung Kim${}^{2*}$
\thanks{$^\dagger$This work was supported by the Robotics and AI (RAI) Institute and Ministry of Trade, Industry \& Energy (MOTIE), Korea (No. 1415187329).}%
\thanks{$^{1}$H. Kim is with the Dept. of Future Automotive Mobility, SNU, Seoul, S. Korea {\tt\small hanjun815@snu.ac.kr}}%
\thanks{$^{2}$M. Jung, C. Noh, S. Jung, H. Song, W. Yang, H. Jang and A. Kim are with the Dept. of Mechanical Engineering, SNU, Seoul, S. Korea {\tt\small [moonshot, gch06208, dan0130, hun1021405, yellowish, dortz, ayoungk]@snu.ac.kr}}
}
\begin{document}


\makeatletter
  \let\@oldmaketitle\@maketitle
  \renewcommand{\@maketitle}{\@oldmaketitle
  \bigskip
  \centering
    \includegraphics[trim= 0cm 0.9cm 0cm 1.2cm, clip,width=0.93\textwidth]{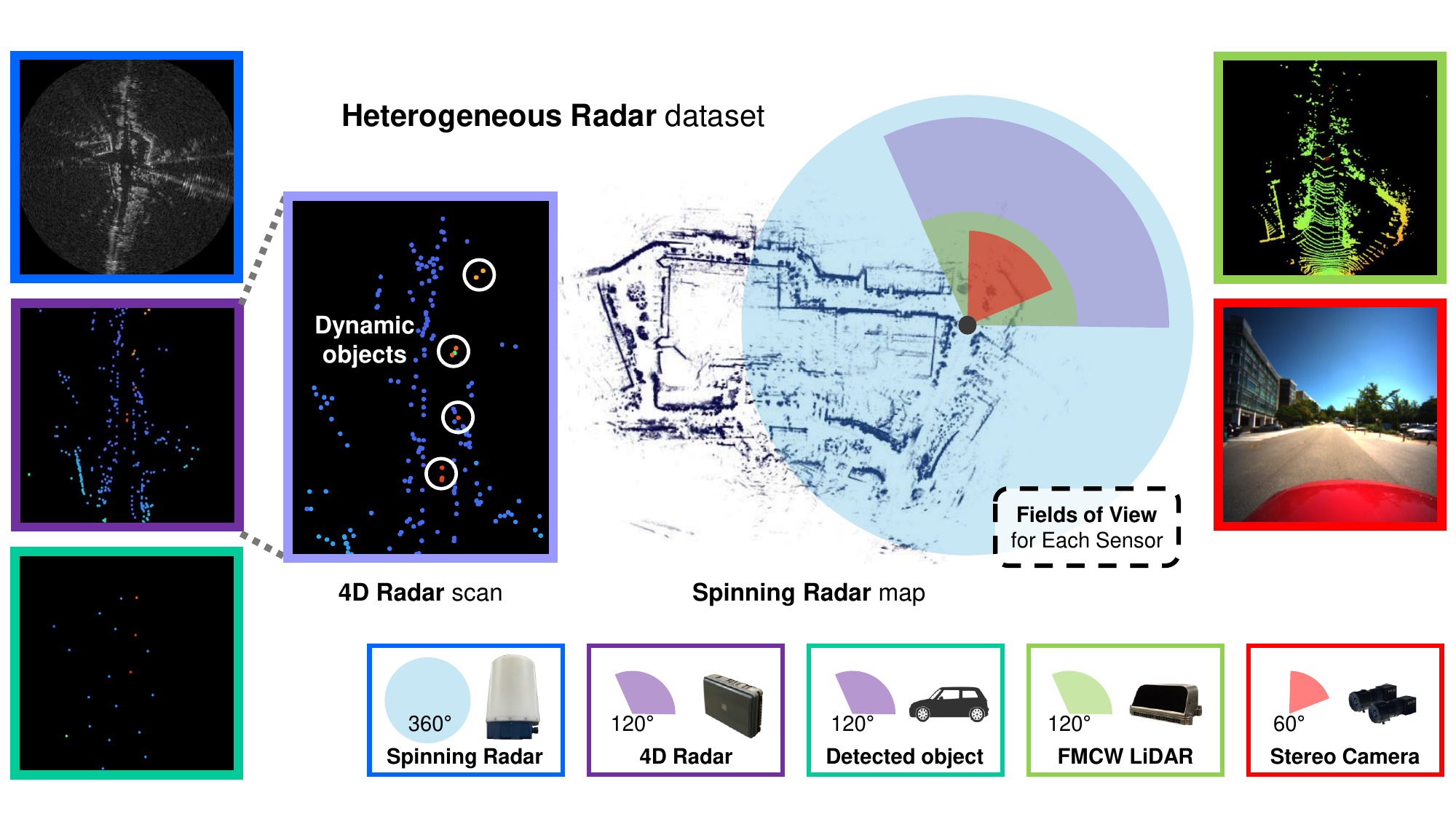}
    \captionof{figure}{
      Overview of the HeRCULES Dataset. The FMCW LiDAR and 4D radar point colors represent relative velocities, with red indicating objects moving away and blue indicating objects approaching. Colors are normalized for each image to enhance visibility.
    }
    \vspace{-3mm}
    \label{fig:main}
  }

\makeatother
\maketitle
\thispagestyle{empty}
\pagestyle{empty}
\setcounter{figure}{1}
\begin{abstract}

Recently, radars have been widely featured in robotics for their robustness in challenging weather conditions. Two commonly used radar types are spinning radars and phased-array radars, each offering distinct sensor characteristics.
Existing datasets typically feature only a single type of radar, leading to the development of algorithms limited to that specific kind. In this work, we highlight that combining different radar types offers complementary advantages, which can be leveraged through a heterogeneous radar dataset. Moreover, this new dataset fosters research in multi-session and multi-robot scenarios where robots are equipped with different types of radars.
In this context, we introduce the HeRCULES dataset, a comprehensive, multi-modal dataset with heterogeneous radars, \acs{FMCW} \acs{LiDAR}, \acs{IMU}, \acs{GPS}, and cameras. This is the first dataset to integrate 4D radar and spinning radar alongside \acs{FMCW} \acs{LiDAR}, offering unparalleled localization, mapping, and place recognition capabilities. The dataset covers diverse weather and lighting conditions and a range of urban traffic scenarios, enabling a comprehensive analysis across various environments. The sequence paths with multiple revisits and ground truth pose for each sensor enhance its suitability for place recognition research. We expect the HeRCULES dataset to facilitate odometry, mapping, place recognition, and sensor fusion research. The dataset and development tools are available at 
\href{https://sites.google.com/view/herculesdataset}{https://sites.google.com/view/herculesdataset}.


\end{abstract}

\section{Introduction}
\label{sec:intro}
Recently, radar has gained significant attention for its reliable performance in conditions such as fog, rain, and low-light environments. Consequently, various radars with different operating modes and unique characteristics have been introduced \cite{harlow2023new}. For example, spinning radar, also known as scanning or imaging radar, offers 360° coverage, a longer perceptible range, and is more resistant to occlusion, making it effective for place recognition or odometry estimation \cite{gadd2020look, park2020pharao, jang2023raplace}. Another widely utilized radar in robotics, phased-array radar, also referred to as \ac{SoC} radar, is lightweight and consumes less power, making it ideal for object tracking in autonomous vehicles \cite{xia2021learning, sengupta2022robust, pearce2023multi}. More recently, 4D \ac{FMCW} radar, which provides elevation information in addition to azimuth and range, has been widely adopted in object detection and \ac{SLAM} \cite{tan20223, cao2018extended, zhuang20234d, zhang20234dradarslam, li20234d}. 

\begin{table*}[h!]
\caption{COMPARISON WITH EXISTING RADAR DATASETS}
\label{table:related}
\resizebox{\textwidth}{!}{%
\begin{tabular}{c|c|cccccccccc} \toprule \rule{0pt}{2ex}
\multirow{2}{*}{\begin{tabular}[c]{@{}c@{}}Radar\end{tabular}} & 
\multirow{2}{*}{Dataset} & 
\multirow{2}{*}{Camera} & 
\multicolumn{2}{c}{Radar} & 
\multirow{2}{*}{LiDAR} & 
\multirow{2}{*}{IMU} & 
\multirow{2}{*}{GPS} & 
\multirow{2}{*}{Size} & 
\multirow{2}{*}{\begin{tabular}[c]{@{}c@{}}ROS \\ Support\end{tabular}} & 
\multirow{2}{*}{Condition} & 
\multirow{2}{*}{Scenarios} \\ 
\cline{4-5} \rule{0pt}{2.3ex}
 &  &  & 4D Radar & Scanning Radar &  &  &  &  &  &  &  \\ \midrule

\multirow{12}{*}{\begin{tabular}[c]{@{}c@{}}4D\\Radar\end{tabular}} & 
  \multirow{-1}{*}{Astyx \cite{8904734}} &
  Mono&
  Astyx 6455 HiRes &
  - &
  3D&  
  - &
  - &
  small &
  - &
  - &
  suburban \\
 &
  \multirow{-1}{*}{RADIal \cite{Rebut_2022_CVPR}} &
  Mono &
  Valeo DDM &
  - &
  3D &
  - &
  RTK &
  medium &
  - &
  - &
  urban, rural \\
 &
  \multirow{-1}{*}{View-of-Delft \cite{palffy2022multi}} &
  Stereo &
  ZF FRGen21 &
  - &
  3D &  
  \checkmark &
  RTK &
  medium &
  - &
  - &
  urban \\
 &
  \multirow{-1}{*}{TJ4DRadSet \cite{zheng2022tj4dradset}} &
  Mono &
  Oculii Eagle &
  - &
  3D &  
  - &
  RTK &
  medium &
  \checkmark &
  night &
  urban \\
&
  \multirow{-1}{*}{K-Radar \cite{paek2022k}} &
  Stereo &
  RETINA-4ST &
  - &
  3D&  
  \checkmark &
  RTK &
  large &
  - &
  {\begin{tabular}[c]{@{}c@{}} night, fog, \\  rain, snow\end{tabular}} &
  {\begin{tabular}[c]{@{}c@{}}urban, suburban, campus,\\ mountain, alleyway  \end{tabular}} \\
 &
  \multirow{-1}{*}{MSC-RAD4R \cite{choi2023msc}} &
  Stereo &
  Oculii Eagle &
  - &
  3D&  
  \checkmark &
  RTK &
  medium &
  \checkmark &
  {\begin{tabular}[c]{@{}c@{}} night, smoke, \\  rain\end{tabular}} &
  {\begin{tabular}[c]{@{}c@{}}urban, rural, tunnel,\\ campus, alleyway  \end{tabular}} \\
 &
  \multirow{-1}{*}{NTU4DRadLM \cite{zhang2023ntu4dradlm}} &
  Mono &
  Oculii Eagle &
  - &
  3D&  
  \checkmark &
  RTK &
  medium &
  \checkmark &
  night &
  campus \\
 &
\multirow{-1}{*}{Dual Radar \cite{zhang2023dual}} &
  Mono &
  {\begin{tabular}[c]{@{}c@{}} Continental ARS548, \\  Arbe Phoenix\end{tabular}} &
  - &
  3D&  
  \checkmark &
  - &
  large &
  - &
  {\begin{tabular}[c]{@{}c@{}} night, dusk, \\ rain\end{tabular}} &
  urban, tunnel \\
 &
\multirow{-1}{*}{Snail radar \cite{huai2024snail}} &
  Stereo &
  {\begin{tabular}[c]{@{}c@{}} Continental ARS548, \\  Oculii Eagle\end{tabular}} &
  - &
  3D&  
  \checkmark &
  RTK &
  large &
  \checkmark &
  {\begin{tabular}[c]{@{}c@{}} night, dusk, \\ rain\end{tabular}} &
  {\begin{tabular}[c]{@{}c@{}} campus, highway, \\  tunnels, overpass\end{tabular}} \\
  \midrule
\multirow{7}{*}{\begin{tabular}[c]{@{}c@{}}Scanning\\Radar\end{tabular}} & 
  \multirow{-1}{*}{Oxford Radar \cite{barnes2020oxford}} &
  Stereo &
  - &
  Navtech CTS350-X &
  3D&  
  - &
  GPS &
  large &
  - &
  rain &
  urban \\
\multicolumn{1}{l|}{} &
  \multirow{-1}{*}{MulRan \cite{kim2020mulran}} &
  - &
  - &
  Navtech CIR204-H &
  3D&  
  \checkmark &
  RTK &
  medium &
  \checkmark &
  - &
  urban, tunnel, campus \\
\multicolumn{1}{l|}{} &
  \multirow{-1}{*}{RADIATE \cite{sheeny2021radiate}} &
  Stereo &
  - &
  Navtech CTS350-X &
  3D &  
  \checkmark &
GPS &
  medium &
  - &
  {\begin{tabular}[c]{@{}c@{}} night, fog, \\  rain, snow\end{tabular}} &
  urban, park \\
\multicolumn{1}{l|}{} &
  \multirow{-1}{*}{Boreas \cite{burnett2023boreas}} &
  Mono &
  - &
  Navtech CIR304-H &
  3D&  
  \checkmark &
  RTX &
  large &
  - &
  {\begin{tabular}[c]{@{}c@{}} night, rain, \\  snow\end{tabular}} &
  urban \\
\multicolumn{1}{l|}{} &
  \multirow{-1}{*}{OORD \cite{gadd2024oord}} &
  Mono &
  - &
  Navtech CTS350-X &
  3D&  
  \checkmark &
  GPS &
  medium &
  - &
  night, snow &
  offroad   \\
  \midrule
  {\begin{tabular}[c]{@{}c@{}} Heterogeneous \\  Radar\end{tabular}} &
  \multirow{-1}{*}{\textbf{HeRCULES}} &
  \textbf{Stereo} &
  \textbf{Continental ARS548} &
  \textbf{Navtech RAS6} &
  \textbf{4D}&  
  \checkmark &
  \textbf{RTK} &
  \textbf{large} &
  \textbf{\checkmark} &
  \textbf{{\begin{tabular}[c]{@{}c@{}} night, dusk, \\ rain, snow\end{tabular}}} &
  \textbf{{\begin{tabular}[c]{@{}c@{}}urban, bridge, campus,\\ mountain, stream, alleyway  \end{tabular}}} \\
\bottomrule
\end{tabular}%
}
\end{table*}

Despite these advancements in radar systems, research integrating multiple types of radars remains less explored. While some datasets and studies have utilized multi-radar setups, they have all featured homogeneous radars \cite{9495184, zhang2023dual, huai2024snail}. This highlights a gap in existing resources, particularly compared to research on heterogeneous radar systems.

\begin{figure}[t]
    \centering
    \includegraphics[trim= 1.6cm 6.7cm 3.9cm 2.1cm, clip,width=0.8\linewidth]{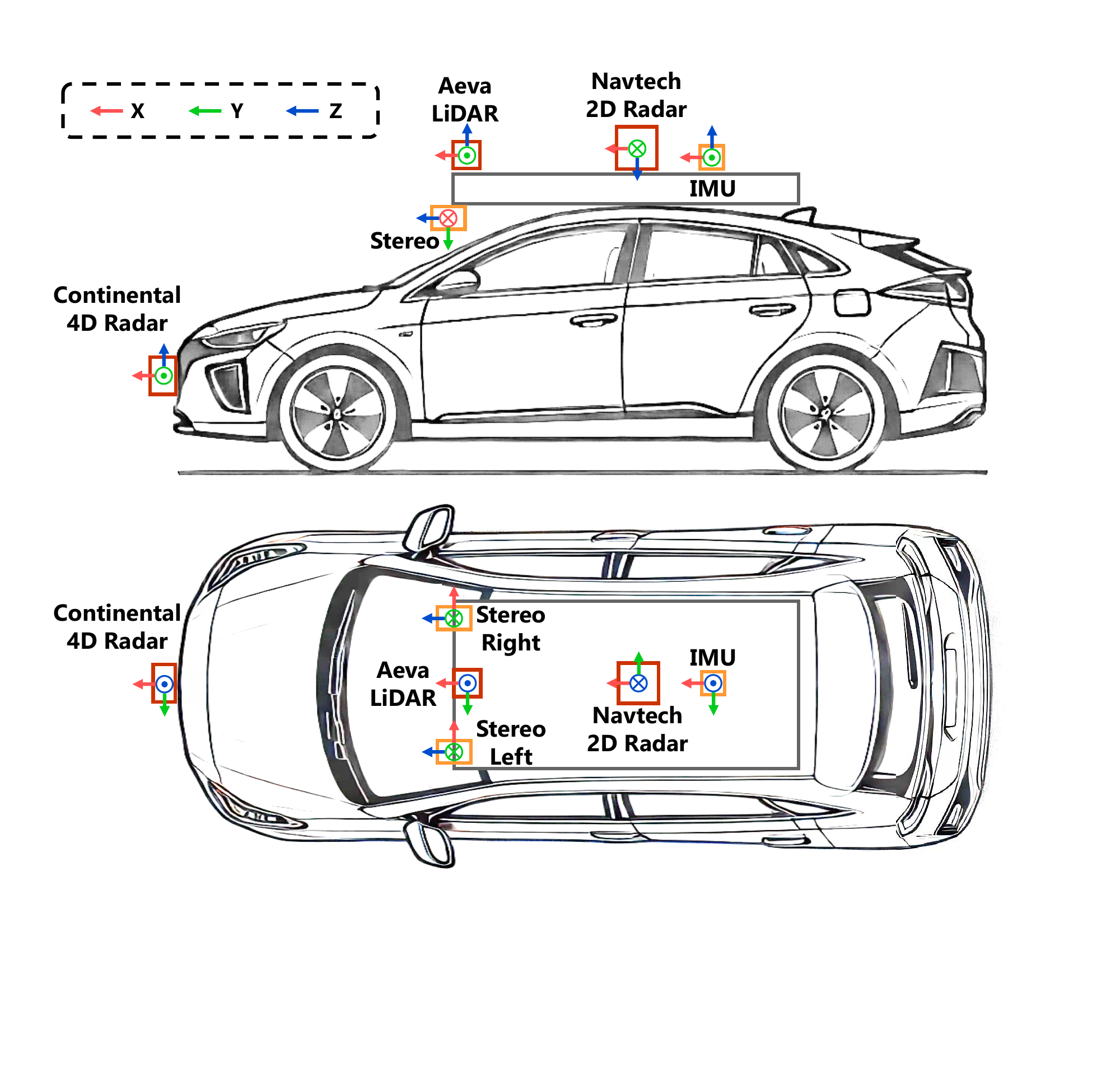}
    \caption{Sensor overview of HeRCULES and coordinate of sensors. The x, y, and z coordinates are red, green, and blue.}
    \label{fig:car}
    \vspace{-5mm}
\end{figure}

To address this gap, we introduce the HeRCULES dataset—\underline{He}terogeneous \underline{R}adar dataset in \underline{C}omplex \underline{U}rban environment for mu\underline{L}ti-s\underline{E}ssion radar \underline{S}LAM—designed to capture rich spatial and velocity information through the combination of heterogeneous radars. This is the first dataset to integrate both 4D radar and spinning radar alongside \ac{FMCW} \ac{LiDAR}, \ac{IMU}, RTK-GPS, and cameras, as shown in \figref{fig:main} and \figref{fig:car}. Instead of using a conventional 3D spinning \ac{LiDAR}, we utilize \ac{FMCW} \ac{LiDAR}, which leverages the advantages of the latest \ac{FMCW} radar by adopting \ac{FMCW} signal methods rather than traditional pulsed laser signals \cite{sayyah2022fully, kim2020fmcw, xu2019fmcw}. This unique setup enables direct comparisons between 4D radar and \ac{FMCW} \ac{LiDAR}, supporting research in radar-\ac{LiDAR} fusion \ac{SLAM} and cross-sensor place recognition. Furthermore, we believe the provided sequences are particularly ideal for multi-session \ac{SLAM} using this heterogeneous sensor setup.

The HeRCULES dataset encompasses various weather and lighting conditions, diverse traffic scenarios, and environments with a large number of dynamic objects. The sequence paths are designed to include multiple revisits to the same locations to support place recognition research. We provide a \ac{ROS} player and radar format conversion software to facilitate easy integration with existing place recognition and SLAM tools. Additionally, the dataset offers ground truth pose for each sensor and presents benchmark evaluations for \ac{SLAM} and place recognition tasks, ensuring comprehensive validation.

Our main contributions are as follows:

\begin{table*}[]
\centering
\caption{SENSOR SPECIFICATIONS}
\label{tab:sensors}
\resizebox{\textwidth}{!}{%
\begin{tabular}{c|cccccccccc} 
\toprule
  \multirow{2}{*}{Sensor} &
  \multirow{2}{*}{Type} & 
  \multirow{2}{*}{Data type} &
  \multicolumn{3}{c}{Resolution}
  &
  &
  \multicolumn{3}{c}{FOV} &
  \multirow{2}{*}{Frequency}
  \\ \cline{4-6} \cline{8-10} \rule{0pt}{2.3ex}
  &
  &
  &
  Range &
  Azimuth &
  Elevation &
  &
  Range &
  Azimuth &
  Elevation &

\\ 
\midrule

  {\begin{tabular}[c]{@{}c@{}}4D\\Radar \end{tabular}} &
Continental ARS548 & 
  {\begin{tabular}[c]{@{}c@{}}x, y, z, velocity, RCS,\\range, azimuth, elavation \end{tabular}} &
\unit{0.22}{m}&
  {\begin{tabular}[c]{@{}c@{}}1.2°@0. . . ±15°\\1.68°@ ±45°\end{tabular}}&
2.3°&
&
\unit{300}{m}&
±60°&
  {\begin{tabular}[c]{@{}c@{}}±4°@\unit{300}{m}\\±14°@$<$\unit{100}{m}\end{tabular}} &
20 Hz
\\ \rule{0pt}{4ex}
  {\begin{tabular}[c]{@{}c@{}}Spinning\\Radar \end{tabular}} &
Navtech RAS6 & 
Polar image, Cartesian image&
\unit{0.044}{m}&
0.9°&
- &
 &
\unit{330}{m} &
360°&
- &
4 Hz
\\ \rule{0pt}{4ex}
  {\begin{tabular}[c]{@{}c@{}}FMCW\\LiDAR \end{tabular}} &
Aeva Aeries II &
  {\begin{tabular}[c]{@{}c@{}}x, y, z, reflectivity, intensity,\\velocity, line-index, time-offset \end{tabular}} &

\unit{0.02}{m}@1$\sigma$ &
0.025°&
0.025°&
&
\unit{150}{m}&
120°&
30° &
10 Hz
\\ \rule{0pt}{4ex}
  Camera &

  {\begin{tabular}[c]{@{}c@{}}FLIR Blackfly S \\BFS-U3-16S2C-CS USB3\end{tabular}} &
8-bit Bayer pattern png format&
- &
\unit{1440}{px} &
\unit{1080}{px} &
&
- &
60°&
45° &
15 Hz
\\ \rule{0pt}{4ex}
IMU & Xsens MTi-300 &
  {\begin{tabular}[c]{@{}c@{}}q$_x$, q$_y$, q$_z$, q$_w$, eul$_x$, eul$_y$, eul$_z$, gyr$_x$, gyr$_y$, \\gyr$_z$, acc$_x$, acc$_y$, acc$_z$, mag$_x$, mag$_y$, mag$_z$\end{tabular}} &
- &
- &
- &
&
- &
- &
- &
100 Hz
\\ \rule{0pt}{4ex}
RTK-GPS & 
  {\begin{tabular}[c]{@{}c@{}}Hexagon NovAtel\\ SPAN-CPT7  \end{tabular}} &
  {\begin{tabular}[c]{@{}c@{}}latitude, longitude, height, velocity$_{north}$,\\velocity$_{east}$, velocity$_{up}$, roll, pitch, azimuth, status \end{tabular}} &
- &
- &
- &
&
- &
- &
- &
50 Hz
\\ 
\bottomrule
\end{tabular}%
}
\vspace{-3mm}
\end{table*}


\begin{itemize}
    \item The HeRCULES dataset is the first public dataset, including both a 4D radar and a scanning radar, providing capabilities for localization, mapping, and place recognition. Moreover, it incorporates the latest FMCW LiDAR with 4D radar, uniquely suited for radar-LiDAR fusion SLAM and cross-sensor place recognition. 
\end{itemize}


\begin{itemize}
    \item The HeRCULES dataset encompasses various weather and lighting conditions, a range of traffic scenarios, and environments with a large number of dynamic objects. The dataset covers an extensive area, enabling comprehensive analysis across various environments.
\end{itemize}

\begin{itemize}
    \item We have designed the sequence paths to include multiple revisits to the same locations. This ensures sufficient queries for place recognition and multi-session \ac{SLAM}.
\end{itemize}


\begin{itemize}
    \item We provide a \ac{ROS} player, radar format conversion software for integration with existing place recognition and SLAM tools, and ground truth poses for each sensor to support place recognition research.
\end{itemize}



\section{Related work}
\label{sec:relatedwork}
\subsection{The Dataset with 4D Radar}
 4D radar captures range, azimuth, elevation, \ac{RCS}, and Doppler velocity, enhancing perception capabilities in dynamic scenarios. While Astyx \cite{8904734} was pioneering in using 4D radar for object detection, its limited diversity and lack of localization references restrict broader applications. Similarly, RADIal \cite{Rebut_2022_CVPR} offers multi-modal data for urban environments but lacks varied environmental conditions and does not include an \ac{IMU}, relying instead on mono camera setups.
View-of-Delft \cite{palffy2022multi} incorporates \ac{GPS}, \ac{IMU}, 4D radar, cameras, and \ac{LiDAR} for object detection and tracking. However, its radar is limited to short-range and lacks long-range 4D radar data. The TJ4DRadSet dataset \cite{zheng2022tj4dradset} focuses on object detection and tracking but excludes adverse weather conditions and lacks an \ac{IMU}. K-Radar \cite{paek2022k} offers a large-scale 4D radar dataset across various weather conditions but only includes a 6-axis \ac{IMU} integrated within the \ac{LiDAR} and lacks radar point cloud data. MSC-RAD4R \cite{choi2023msc} includes stereo cameras, \ac{LiDAR}, RTK-GPS, and \ac{IMU} data over \unit{51.6}{km} but suffers from significant RTK closure errors in height and incorrect headings from the \ac{AHRS} system. NTU4DRadLM \cite{zhang2023ntu4dradlm} is limited in diverse weather conditions, reducing its effectiveness for robust \ac{SLAM} research. Although Dual Radar \cite{zhang2023dual} and Snail Radar \cite{huai2024snail} feature dual radar systems and diverse environments, they only utilize homogeneous radars.

The HeRCULES dataset is the first to combine 4D radar, spinning radar, \ac{FMCW} \ac{LiDAR}, cameras, \ac{IMU}, and RTK-GPS. Unlike other datasets, HeRCULES provides not only the radar point cloud data from the 4D radar but also the object point cloud information filtered through the object filtering process of the ars548 RDI radar driver\footnote{https://github.com/robotics-upo/ars548\_ros/tree/noetic}.

\begin{figure}[!t]
    \centering
    \includegraphics[trim= 10.5cm 3.6cm 11cm 2.85cm, clip,width=1\linewidth]{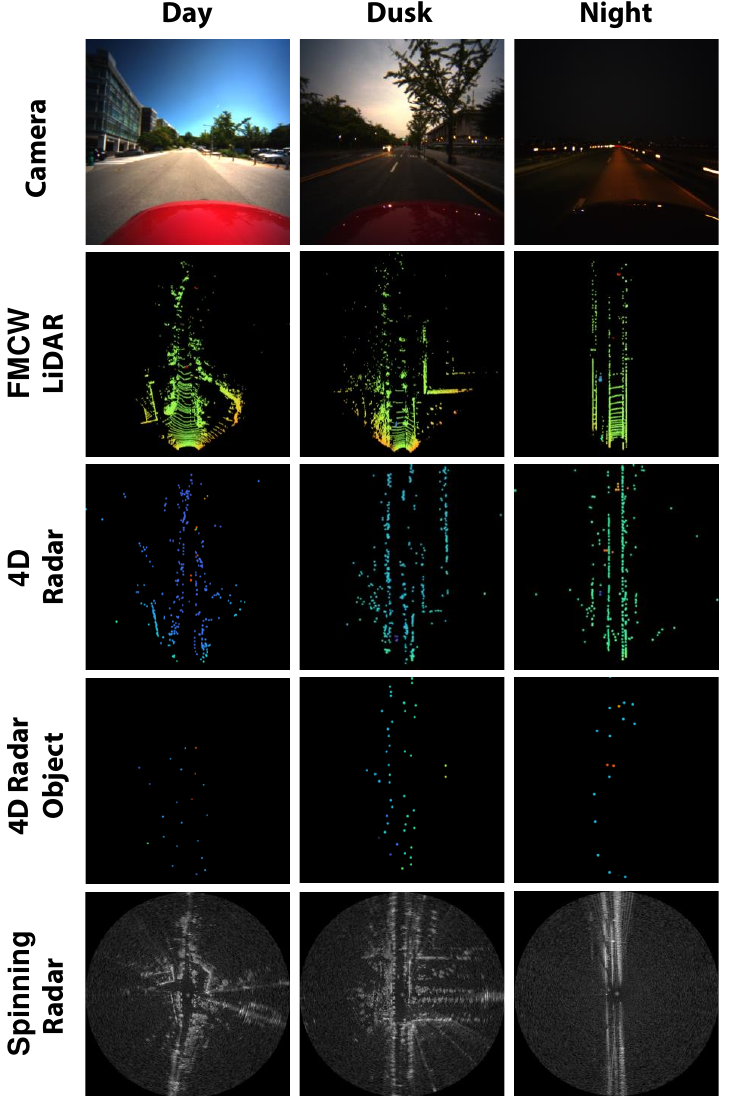}
    \caption{Day, dusk, and night conditions of the HeRCULES dataset.}
    \label{fig:overview}
    \vspace{-8mm}
\end{figure}


\subsection{The Dataset with Spinning Radar}
Spinning radar provides detailed 360° long-range scans, essential for mapping and localization in complex environments. While the Oxford Radar RobotCar Dataset \cite{barnes2020oxford} and MulRan \cite{kim2020mulran} offer valuable data for place recognition and localization in urban settings, they are limited in scope and sensor diversity. The Oxford Radar RobotCar Dataset is restricted to urban areas and lacks scenarios in diverse environments like mountains and river bridges. It also does not include \ac{IMU}, RTK-GPS, and nighttime data. Similarly, MulRan lacks camera data and does not cover varied weather and lighting conditions. The RADIATE \cite{sheeny2021radiate} and Boreas \cite{burnett2023boreas} datasets focus on adverse weather and multi-seasonal environments. However, RADIATE lacks sufficient repeated traversals necessary for robust place recognition tasks. Despite offering a broader range of conditions, Boreas is limited to relatively flat urban and suburban terrains, lacking the complexity of more varied landscapes. The Oxford Offroad Radar Dataset (OORD) \cite{gadd2024oord} features challenging off-road environments but lacks a comprehensive range of urban and rural scenarios.

Compared to existing datasets, the HeRCULES dataset offers several key advantages, as shown in \tabref{table:related}. All the above datasets are limited to 2D radar and do not provide Doppler velocity information. In contrast, HeRCULES combines 4D radar, \ac{FMCW} \ac{LiDAR}, and spinning radar, providing enhanced robustness in SLAM across diverse weather, lighting, urban traffic, and dynamic conditions.

\section{System overview}
\label{sec:overview}
\subsection{System Configuration}



The sensor configuration and coordinates of each sensor are illustrated in \figref{fig:car}, and their specifications are detailed in \tabref{tab:sensors}. The Aeva \ac{LiDAR} operates with relative velocity settings synchronized to the 4D radar sensor. The Continental radar provides both raw radar point cloud data and filtered object point cloud data via its sensor driver. All sensor data are processed on an industrial PC, the NUVO-9006LP-NX, equipped with an Intel Core i9 processor, 2 TB SSD, and 64 GB DDR5 RAM. The sample data is shown in \figref{fig:overview}.





\begin{figure}[!t]
    \centering
    \includegraphics[trim= 7.5cm 17cm 7cm 6.5cm, clip,width=\linewidth]{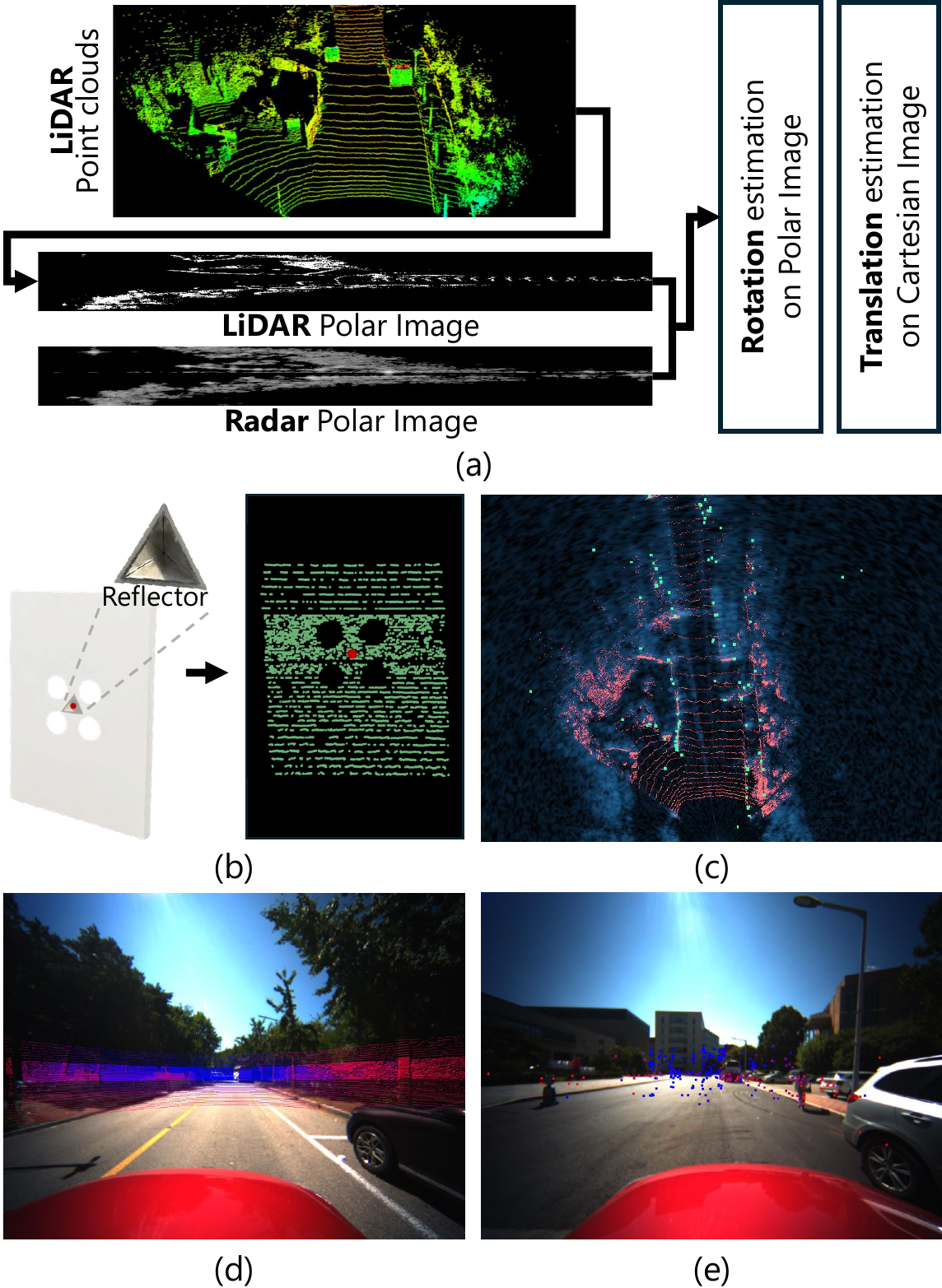}
    \caption{(a) LiDAR - spinning radar extrinsic calibration pipeline. (b) Utilizing the line-index channel. (c) LiDAR points, 4D radar points, and spinning radar points are red, green, and blue. (d)   Right camera - LiDAR. (e) Left camera - 4D radar.}
    \label{fig:cali}
    \vspace{-5mm}
\end{figure}

\subsection{ Sensor Calibration}


\subsubsection{Extrinsic Calibration of \ac{LiDAR} - Spinning Radar}
We employ the method used in the Boreas dataset \cite{burnett2023boreas}. This method determines the rotation $\textbf{R}^L_R$ and the translation $\textbf{t}^L_R$ in the xy plane through correlative scan matching with the Fourier-Mellin transform \cite{checchin2010radar}. Specifically, we convert \ac{LiDAR} point clouds into \ac{LiDAR} polar images to compare with radar polar images to obtain $\textbf{R}^L_R$. Then, we utilize Cartesian images to derive $\textbf{t}^L_R$. To match the \ac{FOV} of the Aeva, we adjust the range and azimuth of the radar images. The calibration pipeline is shown in \figref{fig:cali}(a).

\subsubsection{Extrinsic Calibration of \ac{LiDAR} - 4D Radar - Camera}
We utilize the calibration tool \cite{domhof2021joint} for cameras, \ac{LiDAR}, and radar. This tool jointly calculates relative transformation parameters using a specialized calibration board and reflector. Although we use a solid-state \ac{LiDAR} instead of a spinning one, we utilize the line-index channel to assess laser depth discontinuity, as illustrated in \figref{fig:cali}(b). Unlike \citeauthor{domhof2021joint} \cite{domhof2021joint}, who estimates the reflector position with 2D radar, we can directly obtain the z-value from our 4D radar, resulting in more accurate calibration.

\subsubsection{Extrinsic Calibration of \ac{LiDAR} - \ac{IMU}}
We initialize the system using the method proposed by \citeauthor{zhu2022robust} \cite{zhu2022robust}. This approach was designed for the Livox \ac{LiDAR} series, so it can be seamlessly applied to our solid-state Aeva \ac{LiDAR} without requiring specific targets.

\subsubsection{Calibration Evaluation}
The calibration results are shown in \figref{fig:cali}(c), \figref{fig:cali}(d) and \figref{fig:cali}(e).

\begin{table}[!t]
\caption{OVERVIEW OF SEQUENCES}
\label{tab:sequence}
\centering
\resizebox{\columnwidth}{!}{
\begin{tabular}{c|c|cccccc} 
\toprule
Sequence & Index & Time & Weather & Length & Loop & Target \\ 
\midrule

\multirow{3}{*}{\texttt{Mountain}} & 01 & Day & Clear & \unit{4}{km} & 2 times & \multirow{3}{*}{\begin{tabular}[c]{@{}c@{}}odometry,\\online place recognition,\\global localization \end{tabular}}   \\
 & 02 & Night & Cloud & \unit{4}{km} & 2 times &  \\ 
 & 03 & Day & Snow & \unit{3}{km} & 1.5 times &  \\ 
\midrule

\multirow{3}{*}{\texttt{Library}} & 01 & Day & Clear & \unit{1.6}{km} & 2 times & \multirow{3}{*}{\begin{tabular}[c]{@{}c@{}}odometry,\\online place recognition,\\global localization \end{tabular}}  \\
 & 02 & Night & Cloud & \unit{1.6}{km} & 2 times &  \\ 
  & 03 & Day & Snow & \unit{0.8}{km} & 1 time &  \\ 
\midrule

\multirow{3}{*}{\begin{tabular}[c]{@{}c@{}}\texttt{Sports}\\\texttt{Complex}\end{tabular}}{} & 01 & Day & Clear & \unit{1.4}{km} & 2 times & \multirow{3}{*}{\begin{tabular}[c]{@{}c@{}}odometry,\\online place recognition,\\global localization \end{tabular}} \\
 & 02 & Night & Cloud & \unit{0.7}{km} & 1 time &  \\ 
  & 03 & Day & Snow & \unit{1.4}{km} & 2 times &  \\ 
\midrule

\multirow{4}{*}{\begin{tabular}[c]{@{}c@{}}\texttt{Parking}\\\texttt{Lot}\end{tabular}}{} & 01 &  Day & Clear & \unit{0.4}{km} & \multirow{4}{*}{inter-session} & \multirow{4}{*}{\begin{tabular}[c]{@{}c@{}}odometry,\\global localization \end{tabular}}  \\
 & 02 & Day & Clear & \unit{0.4}{km} &  &  \\
 & 03 & Night & Cloud & \unit{0.5}{km} &  &  \\ 
 & 04 & Day & Snow & \unit{0.4}{km} &  &  \\ 
\midrule

\multirow{3}{*}{\begin{tabular}[c]{@{}c@{}}\texttt{River}\\\texttt{Island}\end{tabular}}{} 
& 01 & Day & Clear & \unit{4}{km} &  & \multirow{3}{*}{\begin{tabular}[c]{@{}c@{}}odometry,\\global localization \end{tabular}}  \\
 & 02 & Dusk & Cloud & \unit{8}{km} & inter-session & \\
 & 03 & Day & Cloud & \unit{5.8}{km} &  &  \\ 
\midrule

\multirow{2}{*}{\texttt{Bridge}} & 01 & Day & Rain & \unit{4.9}{km} & 1 time & odometry, \\
& 02 & Night & Cloud & \unit{4.9}{km} & 1 time & online place recognition \\ 
\midrule

\texttt{Street} & 01 & Day & Rain & \unit{1}{km} & 1 time & odometry \\ 
\midrule

\multirow{2}{*}{\texttt{Stream}} & 01 & Day & Clear & \unit{4.2}{km} & 2 times & odometry, \\

 & 02 & Night & Cloud & \unit{5.5}{km} & 2 times & online place recognition \\ 
\bottomrule
\end{tabular}
}
\vspace{-5mm}
\end{table}


\section{Description of HeRCULES Dataset}
\label{sec:experiment}

\subsection{Target Environments}
\begin{figure}[!t] 
    \centering
    \includegraphics[trim=0.5cm 5.2cm 0.5cm 1.2cm, clip, width=\linewidth]{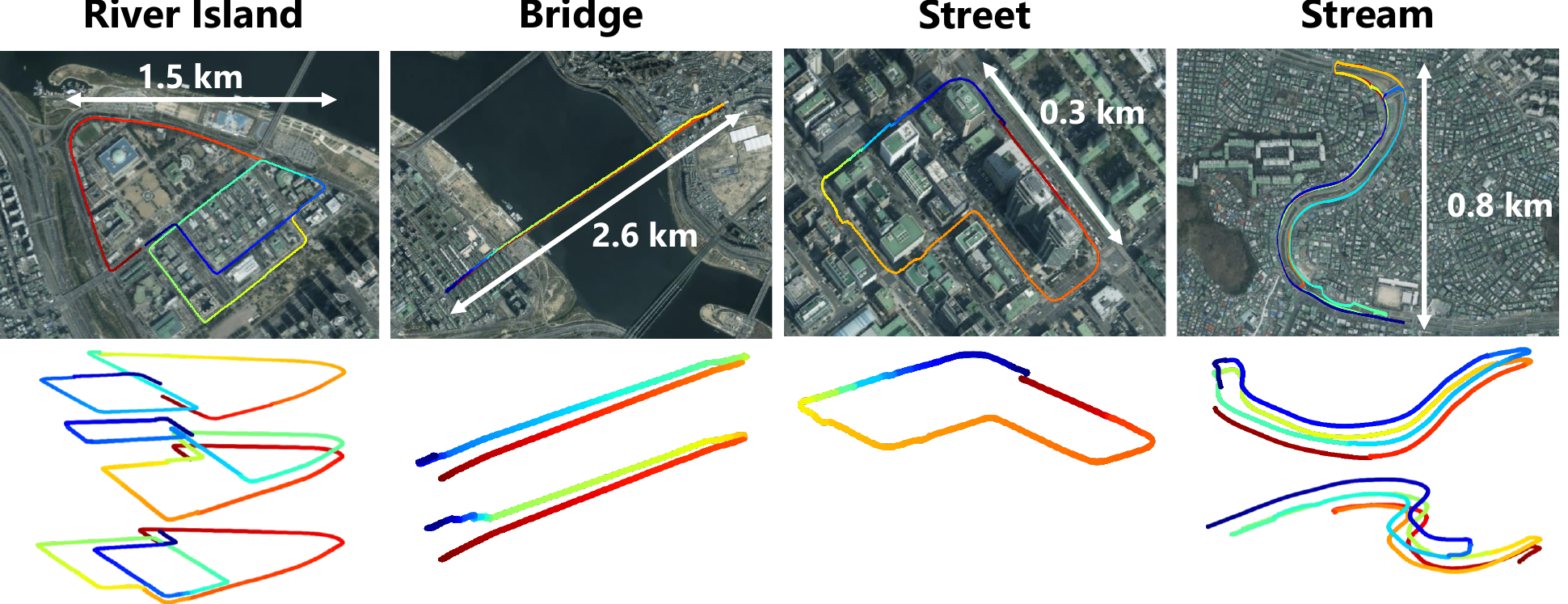} 
    \caption{Trajectory overlaid on satellite maps for each sequence with colors. Red indicates the start, while blue designates the end.}
    \label{fig:road}
    \vspace{-6mm}
\end{figure}

This subsection briefly outlines the reasons for selecting the eight target environments depicted in \figref{fig:road}. An overview of the eight sequences is presented in \tabref{tab:sequence}.

\subsubsection{Mountain}
\texttt{Mountain} captures sequences on Gwanak Mountain, the highest elevation difference among all sequences. The route includes speed bumps and rough roads, causing significant rolling and pitching.

\subsubsection{Library}
\texttt{Library} captures sequences from a long, narrow, one-way path near the library on campus. The path includes steep curves with uphill and downhill sections.
\subsubsection{Sports Complex}
\texttt{Sports Complex} captures sequences around a sports complex, including parking areas and roads with flat, gently sloped, and steep sections. Two loops with an average speed below 30 km/h were recorded during the day and night.
\subsubsection{Parking Lot}
\texttt{Parking Lot} captures sequences from a parking lot with many left turns, recorded on a clear afternoon and at night. While the ground appears flat, slight elevation variations are noted. This sequence has the shortest distance among all.
\subsubsection{River Island}
\texttt{River Island} captures three sequences for multi-session place recognition with each route uniquely designed. The flat area features various driving paths, including intersections and two-lane one-way roads.
\subsubsection{Bridge}
\texttt{Bridge} captures sequences for place recognition research, driven back and forth along a four-lane overpass and the Wonhyo Bridge over the Han River. It includes sequences recorded on a rainy afternoon and cloudy dusk, with an average speed of \unit{60}{km/h}, and sections featuring traffic congestion in urban environments.
\subsubsection{Street}
\texttt{Street} captures a sequence of driving in heavy congestion and rain near IFC Seoul during rush hour. Due to the crowds and numerous vehicles, there are many dynamic objects, leading to frequent stops.
\subsubsection{Stream}
\texttt{Stream} captures an S-shaped stream route with one-way roads running along both sides, allowing U-turns via bridges. For place recognition research, intentional revisits were designed, resulting in similar environments.


\begin{figure}[!t] 
    \centering
    \includegraphics[trim=0.5cm 4.55cm 0.5cm 1.2cm, clip, width=\linewidth]{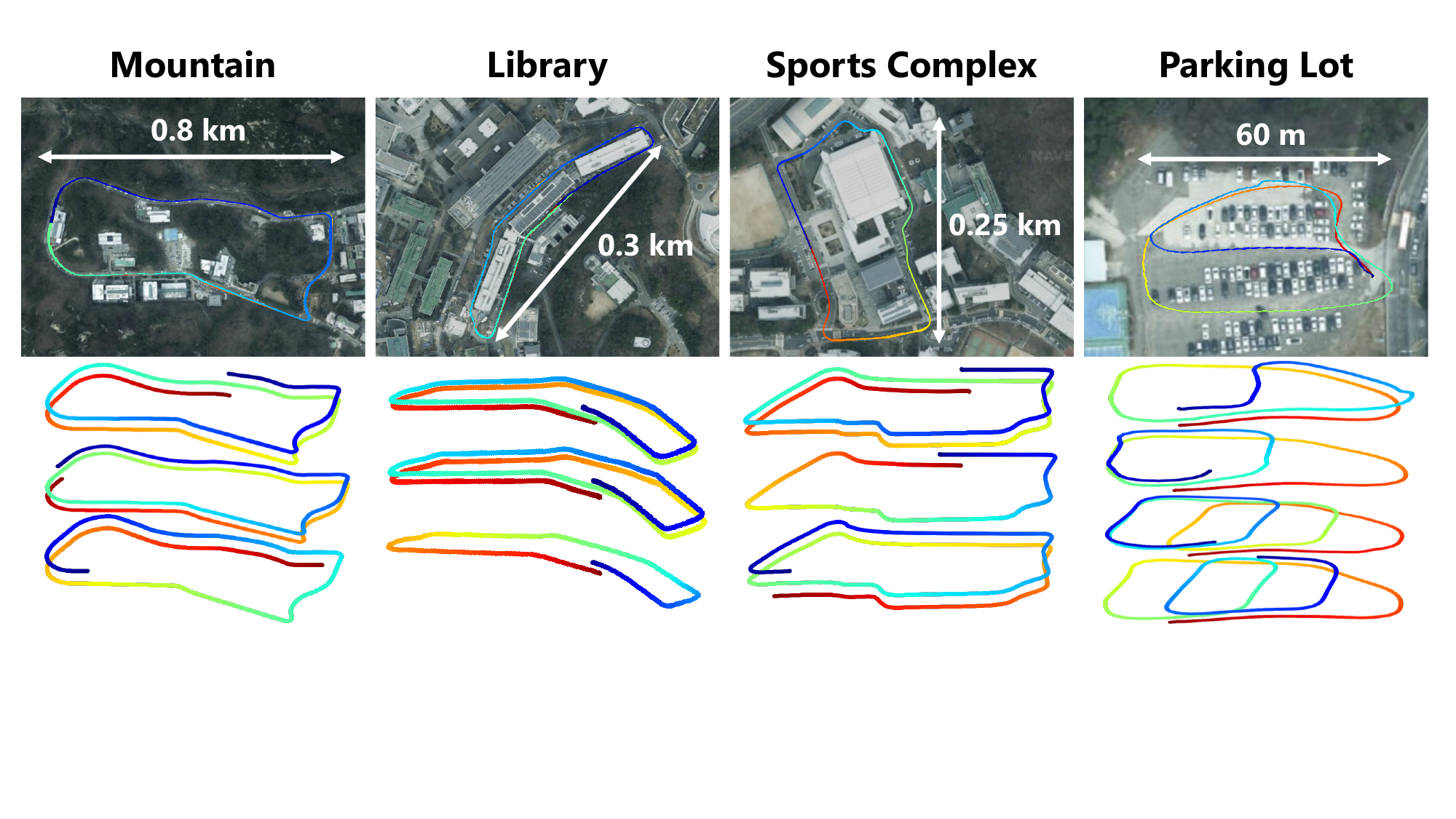} 
    \vspace{-7mm}
\end{figure}

\subsection{Data Description and Format}

\begin{figure}[!t]
    \centering
    \includegraphics[trim= 0cm 1.7cm 0.3cm 1.7cm, clip,width=\linewidth]{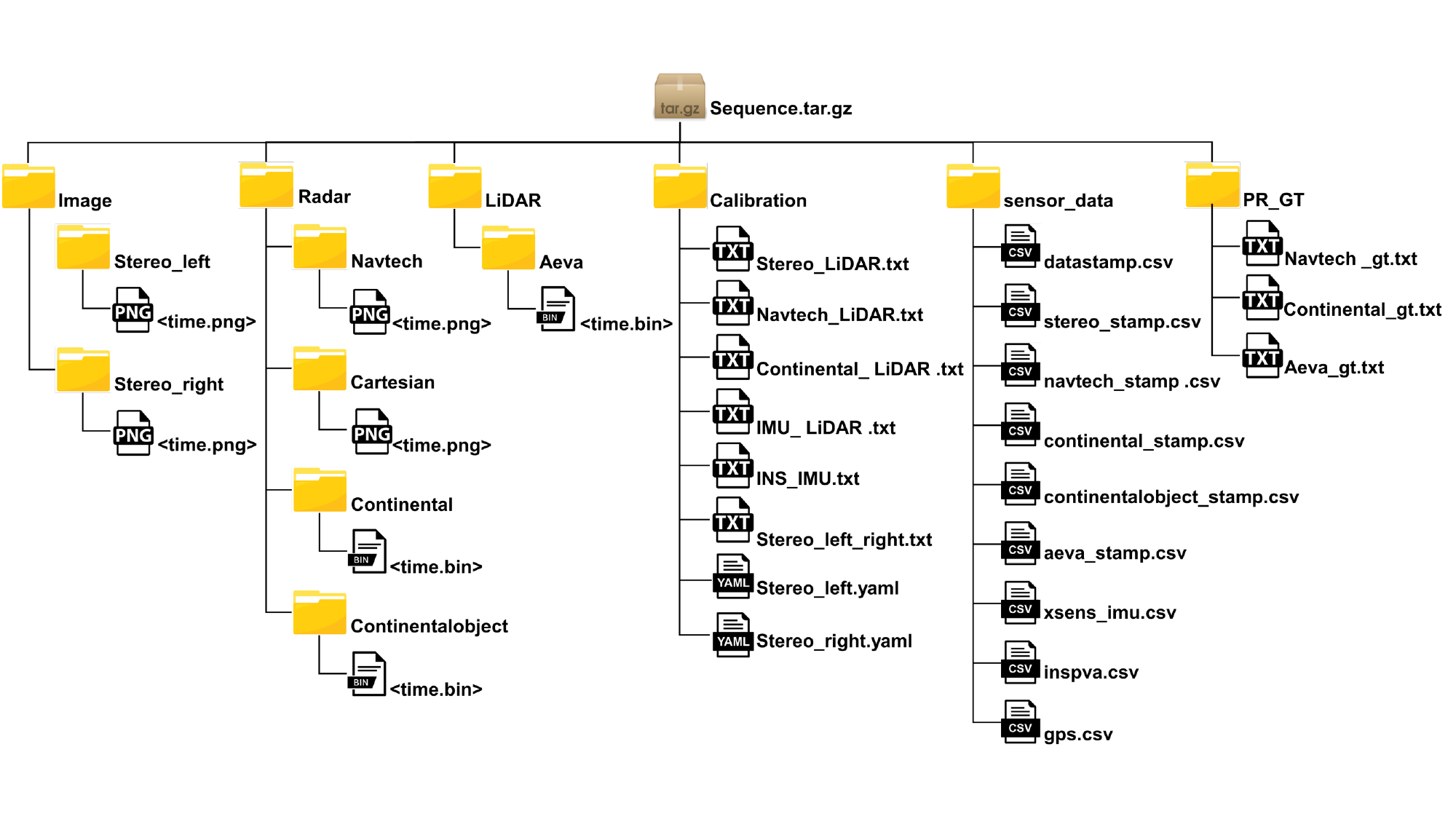}
    \caption{File structure of the HeRCULES dataset, illustrating the organization of sensor scans, ground truths, calibration, and inertial sensor measurements for each sequence.}
    \label{fig:file}
    \vspace{-4mm}
\end{figure}

The file structure of the HeRCULES dataset is delineated in \figref{fig:file}. The acquisition time of
all measurements are stored in \texttt{datastamp.csv}. The FMCW LiDAR and 4D radar data are provided in \texttt{time.bin}, while the camera data is in \texttt{time.png}. For spinning radar data, we support software that converts raw polar images into Oxford-style \cite{barnes2020oxford} and Cartesian images. \ac{IMU}, \ac{GPS}, and \ac{INS} data are provided in \texttt{.csv}, and calibration information between sensors is available in \texttt{.yaml} and \texttt{.txt} format. The data types for each sensor are detailed in \tabref{tab:sensors}.

\subsection{Individual Ground Truth}
\begin{figure}[!t]
    \centering
    \includegraphics[trim= 3.2cm 8.6cm 3.3cm 9cm, clip,width=0.91\linewidth]{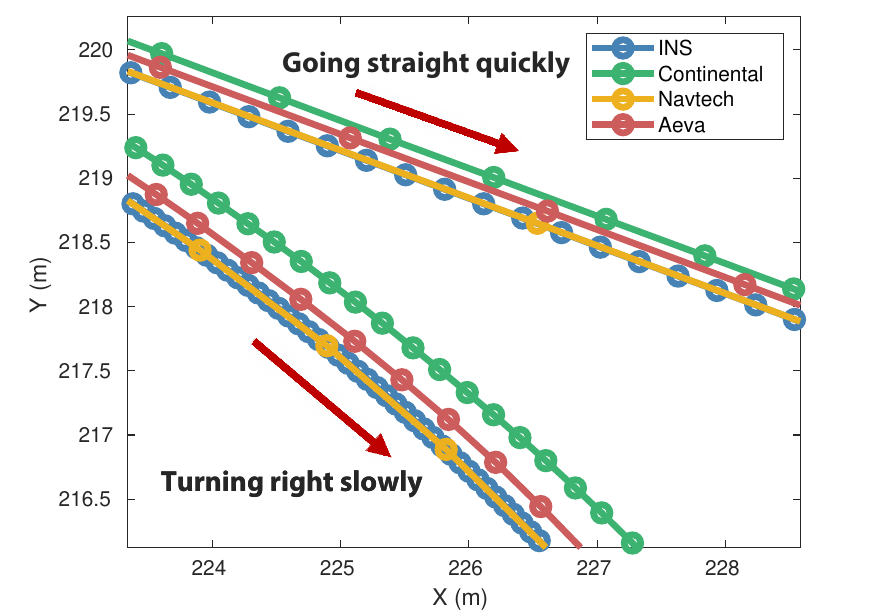}
    \caption{Ground truth pose for each sensor is overlaid on the path of the \texttt{River Island}, illustrating a right turn and a straight drive.}
    \label{fig:gtpose}
    \vspace{-5.8mm}
\end{figure}

Before logging each sequence, we ensure that the GNSS solution is fixed and the INS solution has converged. We use \ac{PTP} to synchronize timestamps in \ac{UTC} across all sensors. However, spatiotemporal discrepancies arise due to differences in sensor mounting positions and data acquisition times.
To address this, we provide ground truth poses for each sensor to support place recognition research, handling spatial differences using extrinsic calibration results and temporal differences with B-Spline interpolation \cite{mueggler2018continuous}.
The ground truth is shown in \figref{fig:gtpose}, highlighting the importance of independently deriving ground truth pose for each sensor.

\section{Evaluation of HeRCULES Dataset}
\label{sec:evaluation}





\subsection{SLAM Evaluation}
\begin{figure}[!t]
    \centering
    \includegraphics[trim= 3.5cm 8.5cm 3.9cm 9cm, clip,width=\linewidth]{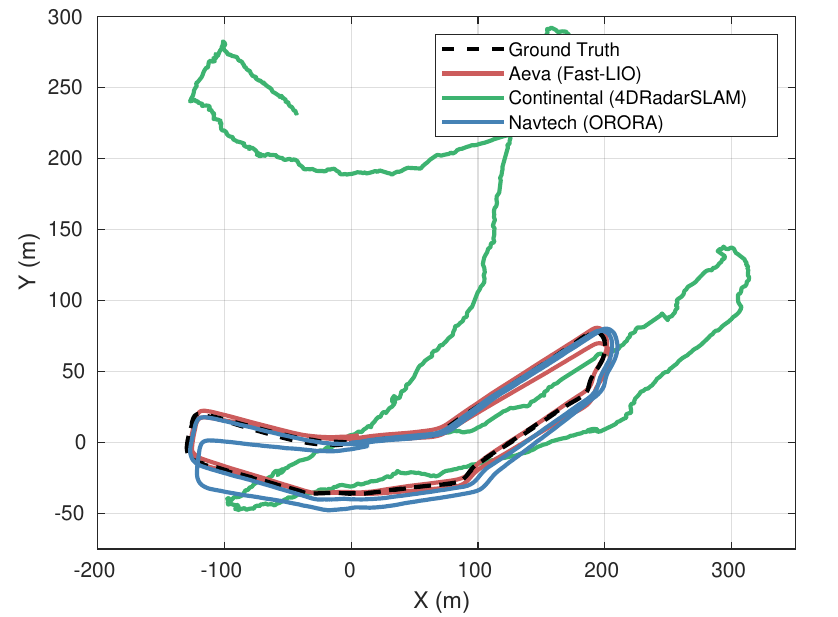}
    \caption{Estimated odometry of Fast-LIO, 4DRadarSLAM and ORORA with the ground truth for \texttt{Sports Complex 01} and \texttt{Library 01}.}
    \label{fig:slam}
    \vspace{-7.8mm}
\end{figure}

For the \ac{SLAM} baseline, we use Fast-LIO \cite{xu2021fast} for \ac{LiDAR} {SLAM}, 4DRadarSLAM \cite{zhang20234dradarslam} for 4D radar \ac{SLAM}, and ORORA \cite{lim2023orora} for spinning radar \ac{SLAM}. \figref{fig:main} shows the mapping results for the \texttt{Sports Complex} using ORORA. A comparison of these three baselines on \texttt{Sports Complex} and \texttt{Library} is shown in  \figref{fig:slam} and  \tabref{tab:slam}. \ac{ATE} is measured in meters, while \ac{RPE} is quantified in degrees per meter for rotation (RPE$_r$) and as a percentage for translation (RPE$_t$). Among these baselines, the odometry result of Fast-LIO was the most accurate, followed by ORORA and 4DRadarSLAM. The result of 4DRadarSLAM is not as good because the point cloud from the Continental radar we used contains fewer points than the Oculii radar originally used in 4DRadarSLAM. However, there is potential for improvement through preprocessing the raw point cloud. These findings validate that \ac{SLAM} performance with 4D radar alone is limited on our dataset, highlighting the need for heterogeneous radar \ac{SLAM} or radar-\ac{LiDAR} fusion \ac{SLAM}.


\begin{table}[t]
\centering
\caption{QUANTITATIVE ANALYSIS: ATE and RPE}
\begin{adjustbox}{width=1\linewidth}
{
\begin{tabular}{c|cccccccccccc}
\toprule
\multirow{2}{*}{Sequence} &
 &
   \multicolumn{3}{c}{Fast-LIO}
 &
 &
  \multicolumn{3}{c}{4DRadarSLAM}
 &
 &
  \multicolumn{3}{c}{ORORA}
  \\ \cline{3-5} \cline{7-9} \cline{11-13} \rule{0pt}{2.5ex}
  &
  &
  ATE$_{t}$ &
  RPE$_r$ &
  RPE$_t$ &
  &
  ATE$_{t}$ &
  RPE$_r$ &
  RPE$_t$ &
  &
  ATE$_{t}$ &
  RPE$_r$ &
  RPE$_t$ 
\\ 
\midrule
\multirow{1}{*}{\texttt{Sports Complex 01}}
& & 
10.358 & 0.950 & 1.763 & &
64.884 & 3.926 & 2.429 & & 
 9.229 & 5.909 & 2.167 \\
\midrule
\multirow{1}{*}{\texttt{Library 01}}
& & 
10.382  & 0.768 & 1.805 &  & 
92.892 & 11.215 &  3.477 & & 
33.348 & 5.143 & 2.045  \\
\bottomrule
\end{tabular}
}
\end{adjustbox}
\label{tab:slam}
\vspace{-2mm}
\end{table}


\begin{figure}[!t]
    \centering
    \includegraphics[trim= 3.6cm 8.5cm 4cm 9.1cm, clip,width=\linewidth]{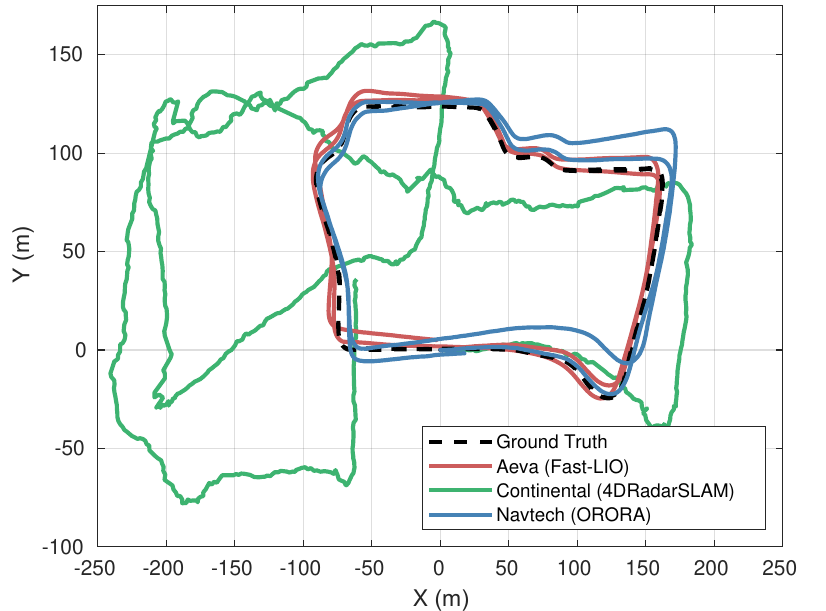}
    \vspace{-8.5mm}
\end{figure}

\subsection{Place Recognition Evaluation}
\begin{figure}[!t]
    \centering
    \includegraphics[trim= 3.9cm 0.8cm 5.7cm 1.8cm, clip,width=\linewidth]{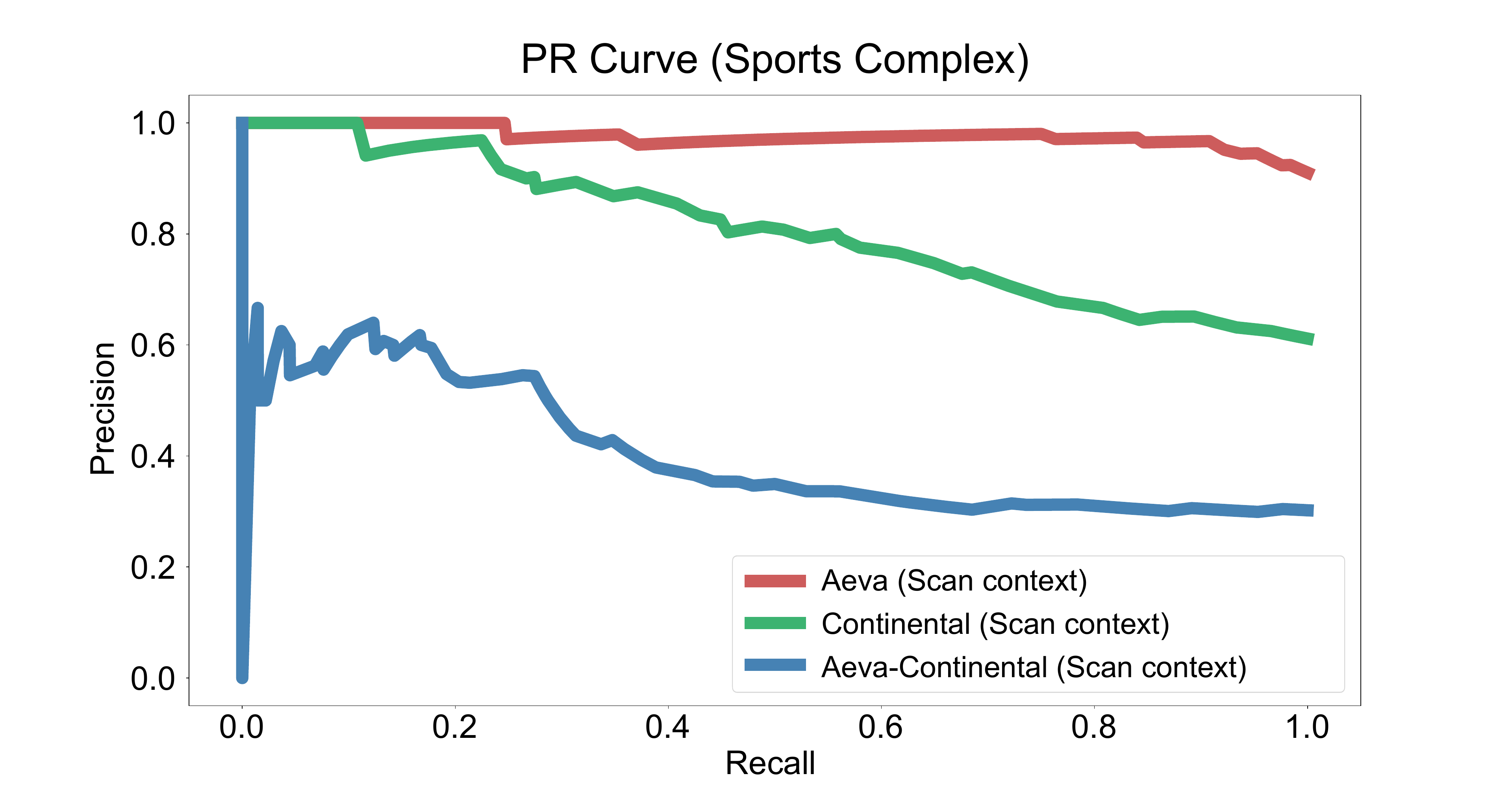}
    \caption{Place recognition result for \texttt{Sports Complex}.}
    \label{fig:pr}
    \vspace{-3mm}
\end{figure}

\begin{figure}[!t]
    \centering
    \includegraphics[trim= 3.9cm 0.8cm 5.7cm 1.8cm, clip,width=\linewidth]{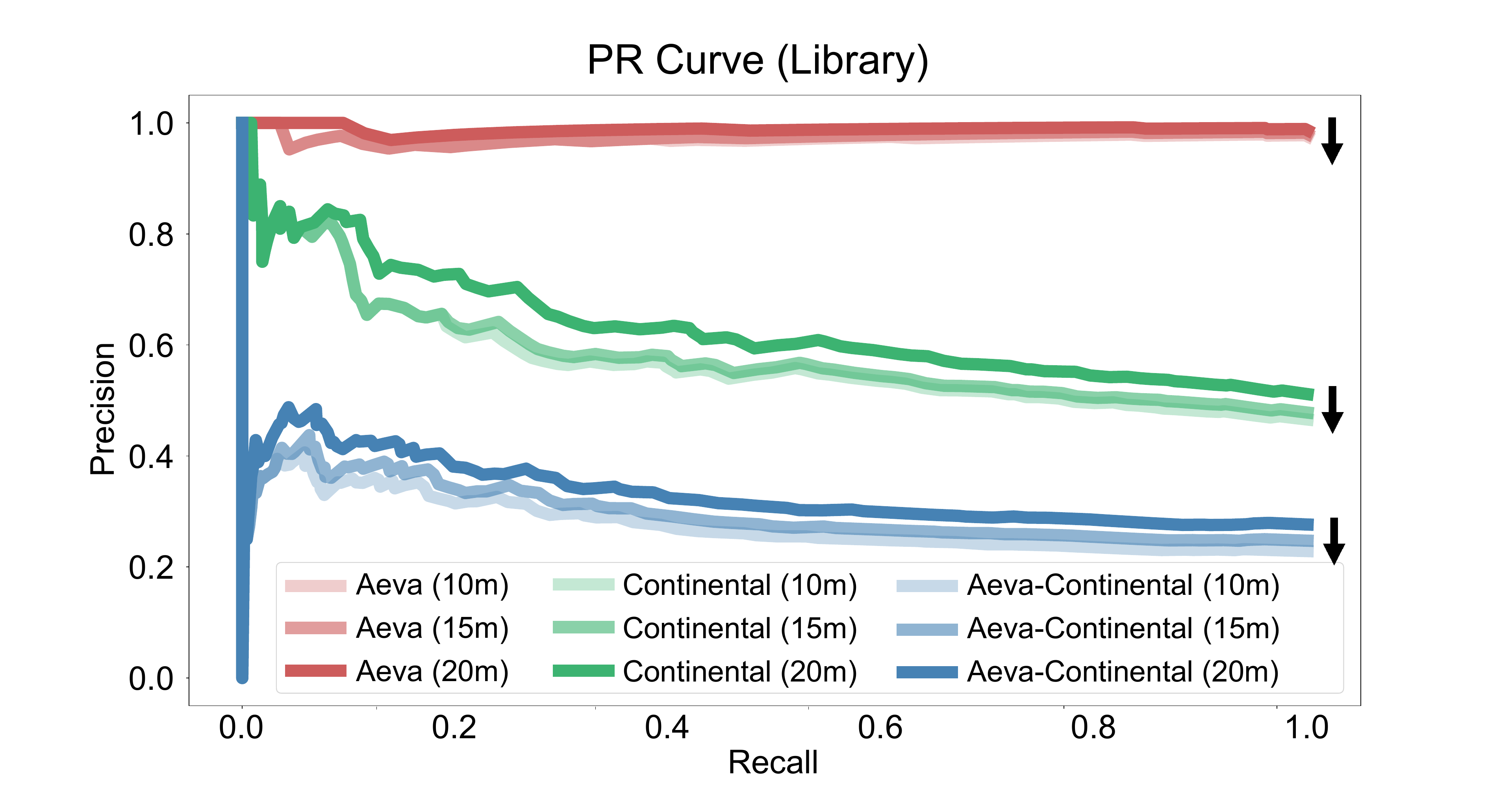}
    \caption{Place recognition result for \texttt{Library}.}
    \label{fig:pr2}
    \vspace{-5mm}
\end{figure}

We use Scan Context \cite{kim2018scan} for place recognition with both 4D \ac{LiDAR} and 4D radar. Additionally, we evaluate cross-modal place recognition performance using radar queries on a \ac{LiDAR}-based database. The experiments utilize \texttt{01} sequences as the database and \texttt{02} sequences as the query set, with the results shown in \figref{fig:pr}. For the \texttt{Sports Complex}, a query is considered correct if the top result is within \unit{10}{m}, indicating accurate recognition of a revisited location or rejection of false positives. The ablation study conducted for the \texttt{Library} shows the results for thresholds of \unit{10}{m}, \unit{15}{m}, and \unit{20}{m} in \figref{fig:pr2}. The AUC scores for the place recognition evaluation are presented in \tabref{tab:pr}.
These results highlight that 4D \ac{LiDAR} achieves the highest place recognition performance, while cross-modal recognition between \ac{LiDAR} and radar is the lowest. This suggests a need for further research in place recognition using heterogeneous sensors, for which HeRCULES offers valuable data.




\begin{table}[t]
\centering
\caption{QUANTITATIVE ANALYSIS: AUC SCORES}
\begin{adjustbox}{width=1\linewidth}
{
\begin{tabular}{c|cccccccccccc}
\toprule
\multirow{2}{*}{Sequence} &
 &
   \multicolumn{3}{c}{Aeva}
 &
 &
  \multicolumn{3}{c}{Continental}
 &
 &
  \multicolumn{3}{c}{Aeva-Continental}
  \\ \cline{3-5} \cline{7-9} \cline{11-13} \rule{0pt}{2.5ex}
  &
  &
  \unit{10}{m} &
  \unit{15}{m} &
  \unit{20}{m} &
  &
  \unit{10}{m} &
  \unit{15}{m} &
  \unit{20}{m} &
  &
  \unit{10}{m} &
  \unit{15}{m} &
  \unit{20}{m}
\\ 
\midrule
\multirow{1}{*}{\texttt{Sports Complex}}
& & 0.976  & -  & -  &   & 0.809 & - & - & & 0.401 & - & -\\
\midrule
\multirow{1}{*}{\texttt{Library}}
& & 0.971 & 0.975 & 0.988   &  & 0.574 & 0.584 & 0.632 & & 0.276 & 0.296 & 0.331 \\
\bottomrule
\end{tabular}
}
\end{adjustbox}
\label{tab:pr}
\vspace{-2mm}
\end{table}


\section{Conclusion}
\label{sec:conclusion}

The HeRCULES dataset is a comprehensive benchmark for \ac{SLAM} and sensor fusion research in autonomous driving, uniquely integrating a diverse sensor suite including 4D radar, spinning radar, \ac{FMCW} \ac{LiDAR}, \ac{IMU}, \ac{GPS}, and cameras. It is the first public dataset to combine 4D radar, spinning radar, and \ac{FMCW} \ac{LiDAR}, offering unique localization, mapping, and place recognition capabilities. Including both 4D radar and \ac{FMCW} \ac{LiDAR} supports diverse research in radar-\ac{LiDAR} fusion \ac{SLAM}, cross-sensor place recognition, and radar point upsampling. Covering diverse weather, lighting, and traffic conditions with sequences that revisit the same locations, the dataset is ideal for robust place recognition and \ac{SLAM} evaluation. Additionally, the HeRCULES dataset includes \ac{ROS}-compatible tools and ground truth pose data for each sensor, facilitating the development of advanced \ac{SLAM} and localization algorithms. Through benchmark evaluations for \ac{SLAM}, we identified the limitations of single radar \ac{SLAM} in various environments, underscoring the need for sensor fusion SLAM research. Additionally, benchmark evaluations for place recognition tasks highlighted the necessity for cross-sensor place recognition research. By offering rich, multi-modal data under varied conditions, HeRCULES sets a new standard for research in autonomous navigation, enabling the creation of next-generation perception systems.




\newpage


\balance
\small
\bibliographystyle{IEEEtranN} 
\bibliography{string-short,references}

\end{document}